\documentclass[10pt,letterpaper,twocolumn]{article}
\usepackage[T1]{fontenc}
\usepackage{lmodern}
\usepackage[letterpaper,top=0.72in,bottom=0.74in,left=0.76in,right=0.76in,columnsep=0.26in]{geometry}
\usepackage{microtype}
\usepackage{graphicx}
\usepackage{booktabs,tabularx,array}
\usepackage[round,authoryear]{natbib}
\usepackage[font=small,labelfont=bf]{caption}
\usepackage{placeins,flushend}
\usepackage[hidelinks]{hyperref}
\hypersetup{
  pdftitle={A Gamified Framework for Conflict-Driven Design: Discovering Strategies for Collaborative Resilience with Multi-Agent Learning},
  pdfauthor={Jin Gao},
  pdfkeywords={urban design, multi-agent reinforcement learning, urban resilience, game theory, participatory planning}
}
\newcommand{\figref}[1]{\mbox{Figure~\ref{fig:#1}}}
\newcommand{\publicationdoi}{10.52842/conf.acadia.2025.2.724}

\begin{document}
\twocolumn[{%
\begin{minipage}{\textwidth}
\centering
{\LARGE\bfseries A Gamified Framework for Conflict-Driven Design\par}
\vspace{3pt}
{\large Discovering Strategies for Collaborative Resilience with Multi-Agent Learning\par}
\vspace{9pt}
{\large Jin Gao\par}
\vspace{2pt}
{\small Massachusetts Institute of Technology\enspace\textbar\enspace
\href{mailto:gaojin@mit.edu}{gaojin@mit.edu}\par}
\vspace{7pt}
{\small Author version published in \emph{ACADIA 2025: Computing for Resilience},
vol.~2, pp.~724--732.\par
\href{https://doi.org/\publicationdoi}{doi:\publicationdoi}\par}
\vspace{7pt}
\includegraphics[width=\textwidth]{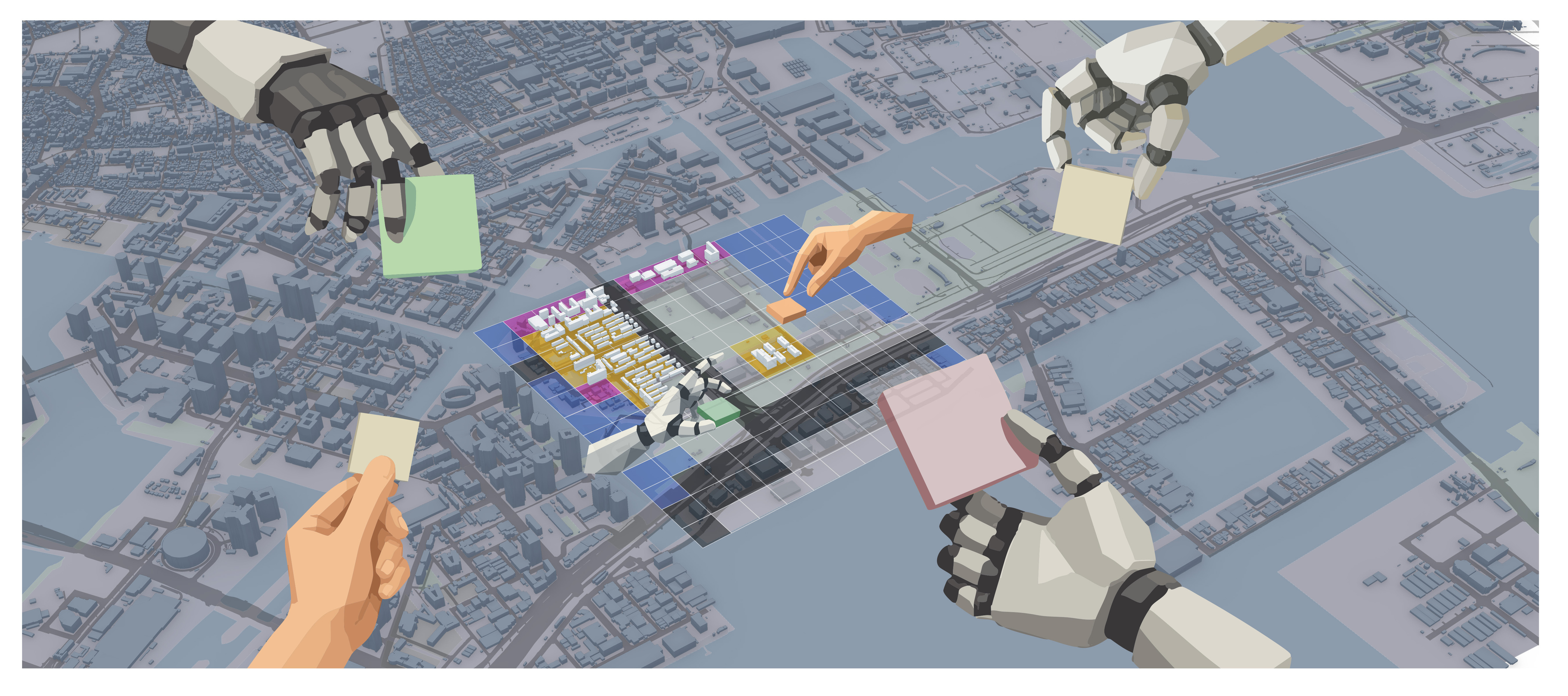}
\captionsetup{hypcap=false}
\captionof{figure}{Illustration of the agent-driven collaborative design process at the Tatsumi case study site.}
\label{fig:1}
\par\vspace{9pt}
\begin{minipage}{0.96\textwidth}
\small
\textbf{Abstract.} Conflicting stakeholder objectives make it difficult to coordinate investments in urban resilience. We model conflict-driven urban design as a non-zero-sum digital board game in which stakeholder agents allocate scarce resources and balance individual payoffs against shared urban development and resilience objectives. We use multi-agent reinforcement learning (MARL) to train policies and examine their behavior across three settings. On a $4\times4$ board, learned policies reach about 84\% of the total return found by a simulated annealing benchmark. On an $8\times8$ board with spatial and resource constraints, trained agents favor projects that improve resilience in self-play. In a $12\times12$ simulation initialized from historical maps of Tatsumi, Tokyo, trained agents achieve a mean total score of 452.1, compared with 357.3 for a random-policy baseline across 10,000 simulation runs. Gains in adaptability, sustainability, flood resistance, and greenery come with reductions in vitality and density. A human-in-the-loop interface supports replay and lets participants take over stakeholder decisions, providing a testbed for exploring these trade-offs in participatory urban planning.

\end{minipage}
\par\vspace{9pt}
\end{minipage}
\vspace{10pt}
}]

\section{Introduction}

In many design projects, one of the most challenging steps is reaching consensus, often shaped by negotiations and exchanges of interests among stakeholders. Our research introduces a customized digital "board game" that provides a minimal abstraction of the real-world city-building process by modeling core stakeholder personas, their roles, and interactions. Within this environment, we apply multi-agent reinforcement learning (MARL) to train AI agents to develop strategies for resource allocation under competing objectives. The trained models are then used to simulate planning processes in conflict-driven scenarios, offering insights into negotiation dynamics and stakeholder trade-offs. Furthermore, we present a human-in-the-loop interface that enables participants to visualize the simulation process and interact with AI agents, supporting a deeper understanding of planning dynamics while offering valuable insights into real-world dilemmas. See \figref{1} for the Tatsumi site.

\section{Related Work}

Board games have long been established as participatory tools. In the board game community, city building is an important genre, exemplified by games such as Bay Area Regional Planner, Small Cities, and Suburbia. Serious games are also widely used in academic research \citep{poplin2012,sandbhor2021}, where interactive interfaces have emerged for participatory planning, bridging the gap between algorithmic optimization and public participation, such as CityScope \citep{alonso2018}.

Game theory provides theoretical foundations in modeling and evaluating the decision process and equilibria. Prior studies have applied it to model land-use negotiations among stakeholders \citep{liu2016}, and collective decision-making processes for urban planning \citep{abolhasani2022}. These studies provide theoretical foundations for analyzing interdependent decision-making in urban contexts.

Agent-based models (ABMs) have been widely used to capture population-level interactions and emergent behaviors in urban environments \citep{batty2005,bijandi2021,crooks2008}. However, ABM agents are often built on simplified rule-based behaviors that have limited capabilities for making complex decisions.

Multi-agent reinforcement learning offers a computational extension to these models, where agents learn strategies through self-play and train neural networks based on interactions in a customized environment. \citet{kaviari2019} combined agent-based modeling and game theory to simulate urban growth, while \citet{qian2023} used MARL to model participatory planning under competing stakeholder interests. Cicero demonstrated how strategic reasoning and language models can be combined in the negotiation game Diplomacy \citep{metafair2022}.

Our research contributes to the realm by proposing a scalable methodology that abstracts the urban decision-making process via games and uses a customized environment to perform efficient MARL training. Agents are trained and evaluated under different reward matrices to study how value propositions affect strategy cooperation under real-world contexts. The workflow contributes to the emerging field of multi-agent learning and computational urban planning by offering real-world use cases and a gamified testbed for collaborative decision-making. Our vision of this project is to shift the traditionally top-down urbanization process into an AI-augmented, participatory approach, empowering diverse voices to be captured with players' reward functions and trained under interpretable approaches.

\section{Methods}

Our method consists of three tightly coupled components. First, we introduce a scalable game environment that simulates core stakeholder interactions. It serves both as a participatory abstraction for modeling real-world planning dynamics and as a testbed for agents to act and learn. Then, we introduce a multi-agent learning workflow, where agents learn adaptive strategies under different value matrices in the game environment. Last, we present a digital human interface to allow humans to observe the game status and interact with agents during the decision process.

\subsection{Game Environment}

Our goal is to define a game environment that captures the complexity of collective decision-making while remaining feasible for agent training. We found that a board game with discrete actions fits this purpose. A central design consideration is the level of abstraction: The game should preserve the key dynamics involved in the design process and balance between stakeholders. For example, when simulating investment-reward dynamics, a critical decision is whether players should continuously receive rewards from their investments over subsequent turns---a factor that directly shapes agents' perspectives on long-term benefits. These nuanced design decisions require fine-tuning to balance between reflecting real-world trade-offs and enabling agents to learn long-term strategies within computational constraints.

\textbf{The game map} consists of a configurable $M\times N$ grid. Each cell has key parameters that represent the environmental, economic, and social aspects of urban development. These parameters are initialized based on the original terrain and any preexisting projects, and are dynamically updated each round according to the buildings placed on the board and their effects on both the selected and adjacent cells. The overall environment score is calculated as the average of all indices across the grid. During each turn, players act sequentially, choosing either to construct a building on an available cell or to skip their turn.

\textbf{Agents} in our framework represent stakeholders whose actions shape the environment. Since stakeholders often pursue diverse and sometimes conflicting objectives, we model their priorities using value matrices, where each player's reward is parameterized by weights that balance self-interest against map parameters representing site conditions and resilience. To capture this diversity, we define an integrated reward function that combines individual resource gains with the collective impacts encoded in the game map parameters.

\begin{figure*}[!t]
\centering
\includegraphics[width=0.93\linewidth]{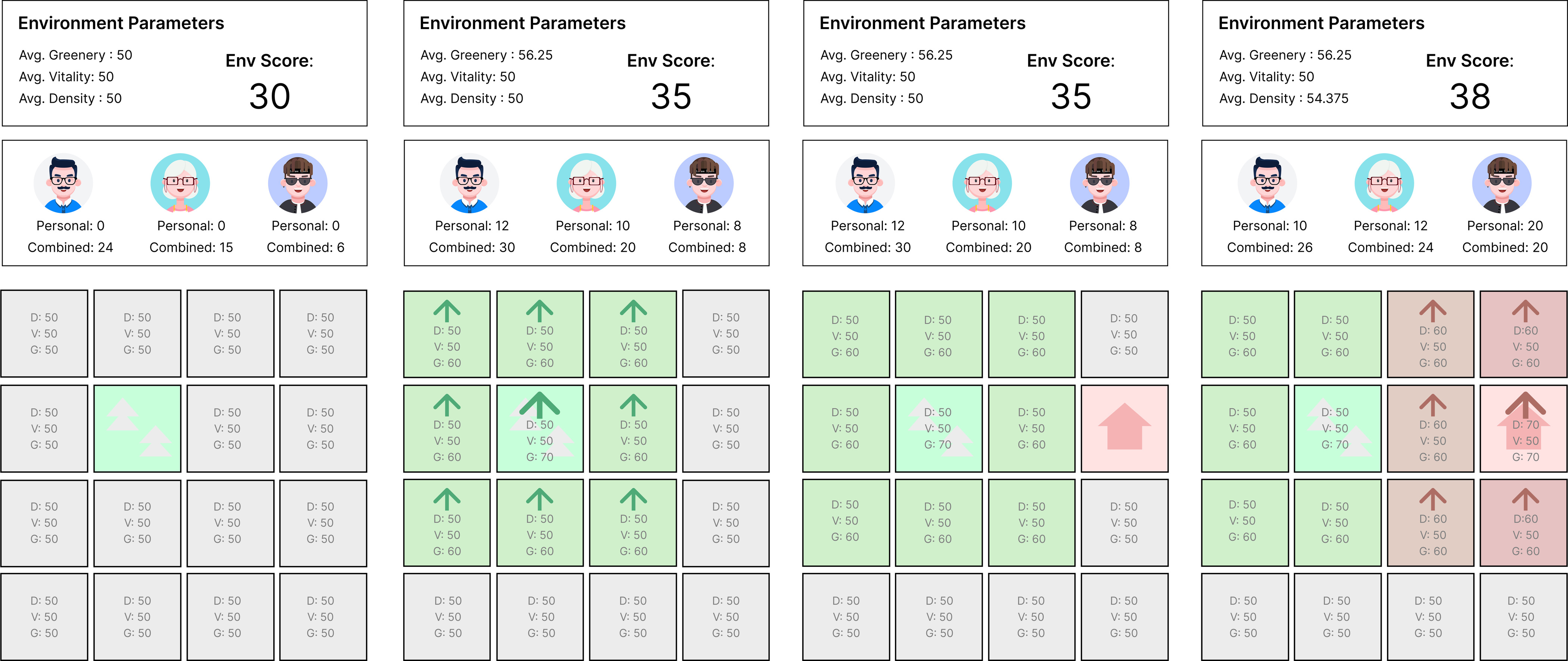}
\caption{Game actions and their effects on player resources.}
\label{fig:2}
\end{figure*}

\textbf{Actions} are categorized into basic and resilience-building projects under the urban resilience theme, each requiring resources to initiate. We define financial capital as the "money" resource, and social capital as the "reputation" resource. A successful action triggers immediate changes on the map and introduces long-term costs and benefits over subsequent turns. For example, constructing a park imposes ongoing maintenance costs on the builder but simultaneously increases their reputation each turn. These trade-offs force players to carefully consider (learn) the short-term and long-term impacts of their actions. In the applied experiment, we further extend the reward mechanism by introducing value decay and a block-level redevelopment rule above a threshold, which allows agents to replace existing buildings with new ones, thereby simulating urban renewal.

\textbf{Interactions} between players, illustrated in \figref{2}, are realized through spatial and temporal dynamics on the game board. In our experiments, constructing a building changes the builder\textquotesingle{}s rewards and the state of the surrounding $3\times3$ neighborhood. These effects are carefully defined to be competitive across building types, so that each building type has distinct characteristics without a single dominant action. For instance, parks increase greenery but decrease vitality, whereas houses reduce greenery while increasing density. This design allows each action to carry distinct trade-offs, encouraging players to strategize based on their individual priorities.

\begin{figure}[!htbp]
\centering
\includegraphics[width=\columnwidth]{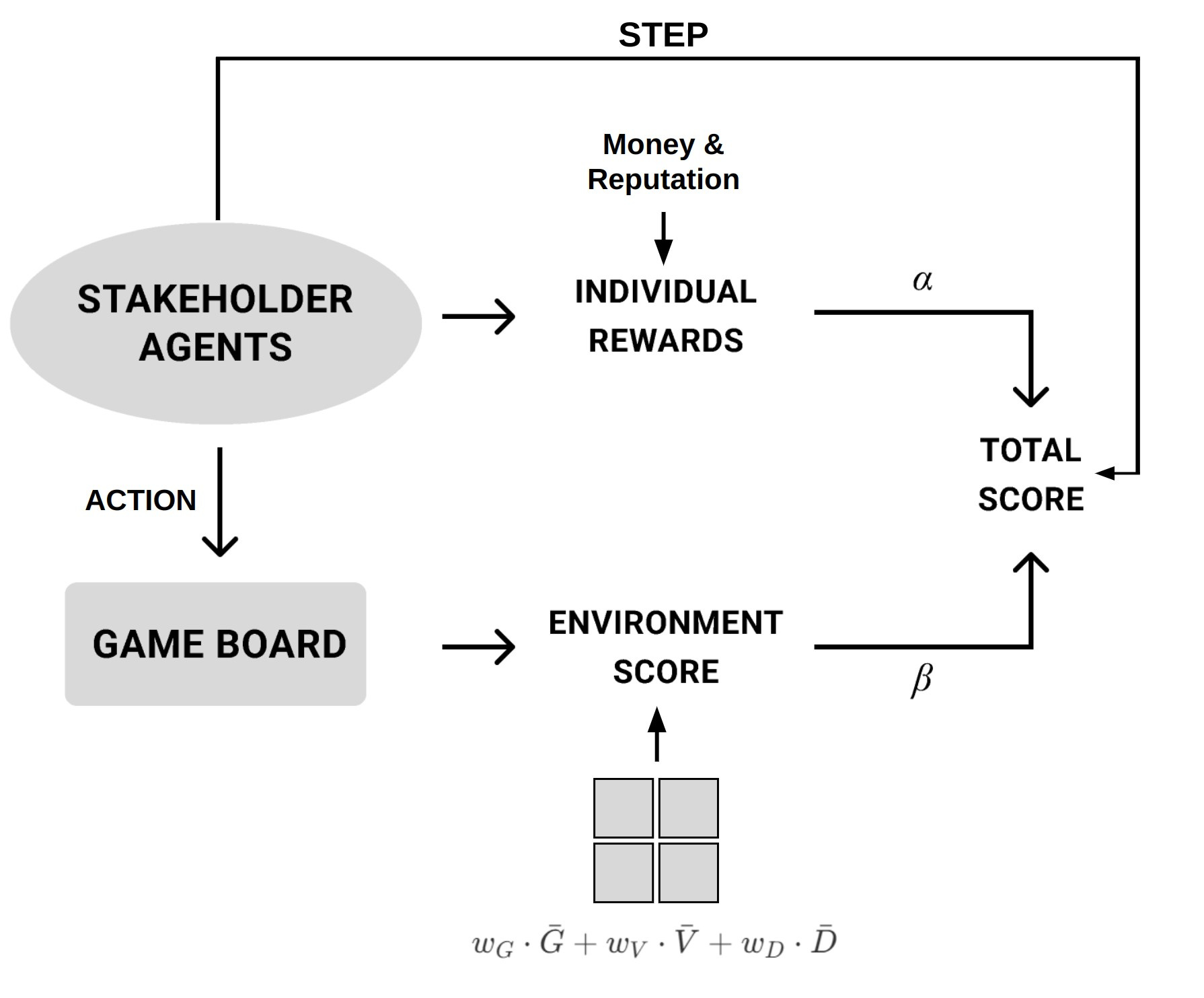}
\caption{The sequential game loop for stakeholder actions and state updates.}
\label{fig:3}
\end{figure}

\textbf{The game loop} is an iterative process where stakeholder agents follow a sequential decision loop to modify the game board. As illustrated in \figref{3}, each action updates both individual rewards and the environmental scores. These components are combined into a total score, parameterized by weights, to capture the balance between personal objectives and collective impact. While no explicit winning criterion is defined, the overall outcome is evaluated collectively: strategies that yield higher total rewards across all agents are considered more desirable. To discourage infeasible actions, players who lack sufficient resources incur a significant penalty.

\subsection{Multi-Agent Learning}

\textbf{Problem Definition:} Unlike well-studied games like poker, tic-tac-toe, or rock-paper-scissors, which are governed by clear rules and often exhibit zero-sum properties, the proposed game environment is non-zero-sum in nature. That is, stakeholders typically have conflicting objectives, but their actions can also generate shared benefits or collective challenges. This makes it challenging to use the properties and theories of zero-sum games to analyze the expected outcome of the game.

Agents learn decision policies from their observations of the environment. Due to the non-zero-sum nature, the players can jointly improve their outcomes by coordinating actions that raise the overall environment score. Consequently, the solution space includes equilibria that are not strictly competitive but rather mutually beneficial. We discuss solution behavior through the lens of coarse correlated equilibrium (CCE), a tractable relaxation of correlated equilibrium (and thus a superset of Nash equilibria) for general-sum games. A CCE represents a strategy distribution under which no agent can improve its expected payoff by committing in advance to a unilateral deviation. It can be approached through learning-based methods. Under such learning dynamics, coordinated strategies may give rise to mutually beneficial outcomes that improve both individual payoffs and the overall environmental score. Less cooperative players, such as interest-driven players, may prioritize short-term gains, which can impede convergence toward coordinated behavior during training.

\textbf{Learning Environment:} While single-agent RL environments are typically defined by a single observation--action--reward loop, multi-agent settings introduce additional challenges such as nonstationarity, coordination, and credit assignment. After comparison, we adopt PettingZoo \citep{terry2021} as the environment API and EPyMARL \citep{papoudakis2021} as the training stack, due to their lightweight design and good interoperability with modern libraries. The core components include the observation space, action space, step function, and reward. The observation space defines what agents perceive at each turn while taking actions. Depending on the setting, agents may receive full or partial observations, which affect the coordination dynamics and learning stability. We define the observation space as a collection of matrices, including the status of the grid, the player's current resources, and the distribution of builders and building types on the map. The action space is discrete and turn-based, where each agent chooses either to construct a building on an available cell or to skip the turn. The step function governs the core game logic and updates the game state. It begins by decoding agents' actions to update environment parameters, then calculates each agent's rewards. Reward design plays a critical role in shaping emergent strategies. In our environment, rewards consist of two primary components: individual rewards, reflecting agents\textquotesingle{} financial resources and social reputation; and an environmental reward, calculated as the average of Greenery (G), Vitality (V), and Density (D) parameters across the grid. We also incorporate an optional resilience reward component alongside traditional metrics, computed as the average of Adaptability (A), Sustainability (S), and Flood Resistance (F) parameters, representing long-term climate adaptation capacity and disaster preparedness indicators. This creates a three-component utility function where agents balance personal resources, traditional urban development metrics, and resilience objectives according to individual preference weights.

\begin{figure*}[!t]
\centering
\includegraphics[width=\linewidth]{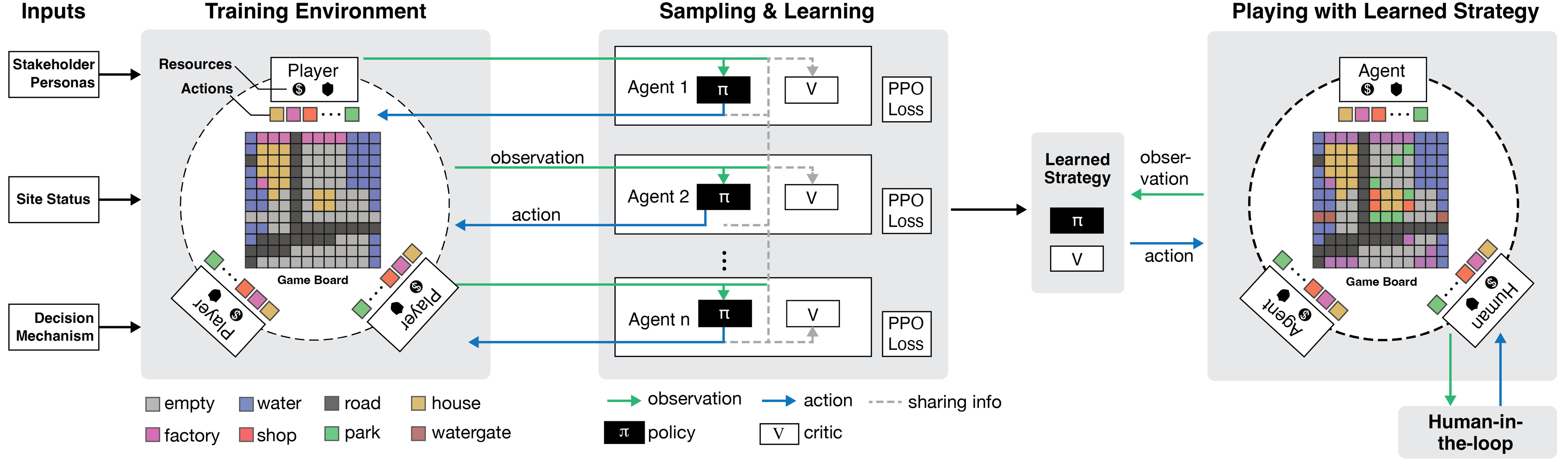}
\caption{Training architecture adapted by the author from the MARLlib proximal policy optimization family \citep{hu2023}. The source is licensed under \href{https://creativecommons.org/licenses/by/4.0/}{CC BY 4.0}.}
\label{fig:4}
\end{figure*}

\begin{figure}[!htbp]
\centering
\includegraphics[width=\columnwidth,trim={0bp 603bp 0bp 0bp},clip]{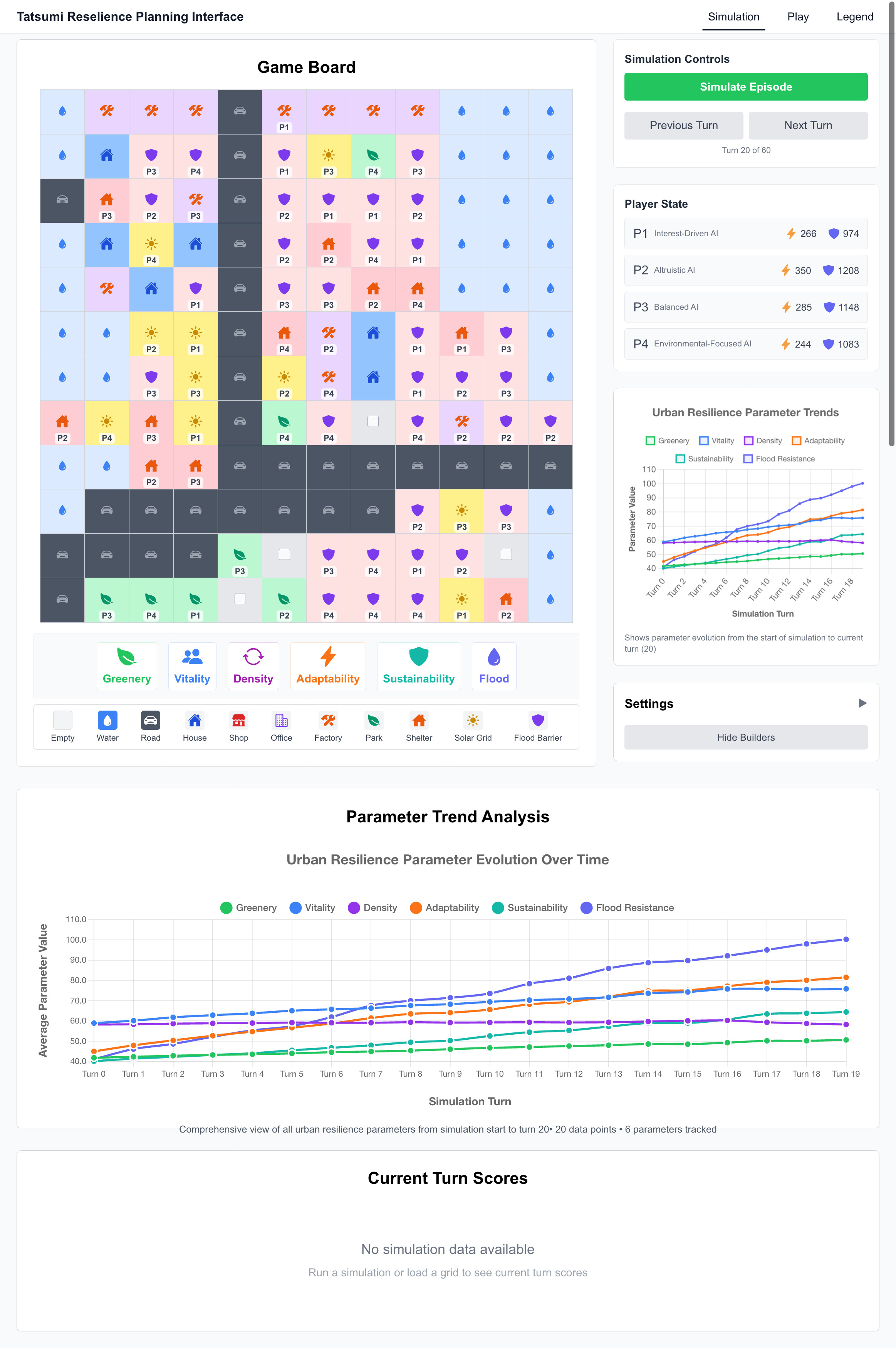}
\caption{Human--agent interface for observing simulations and taking over an agent's decisions.}
\label{fig:5}
\end{figure}

\begin{figure*}[!t]
\centering
\includegraphics[width=0.88\linewidth]{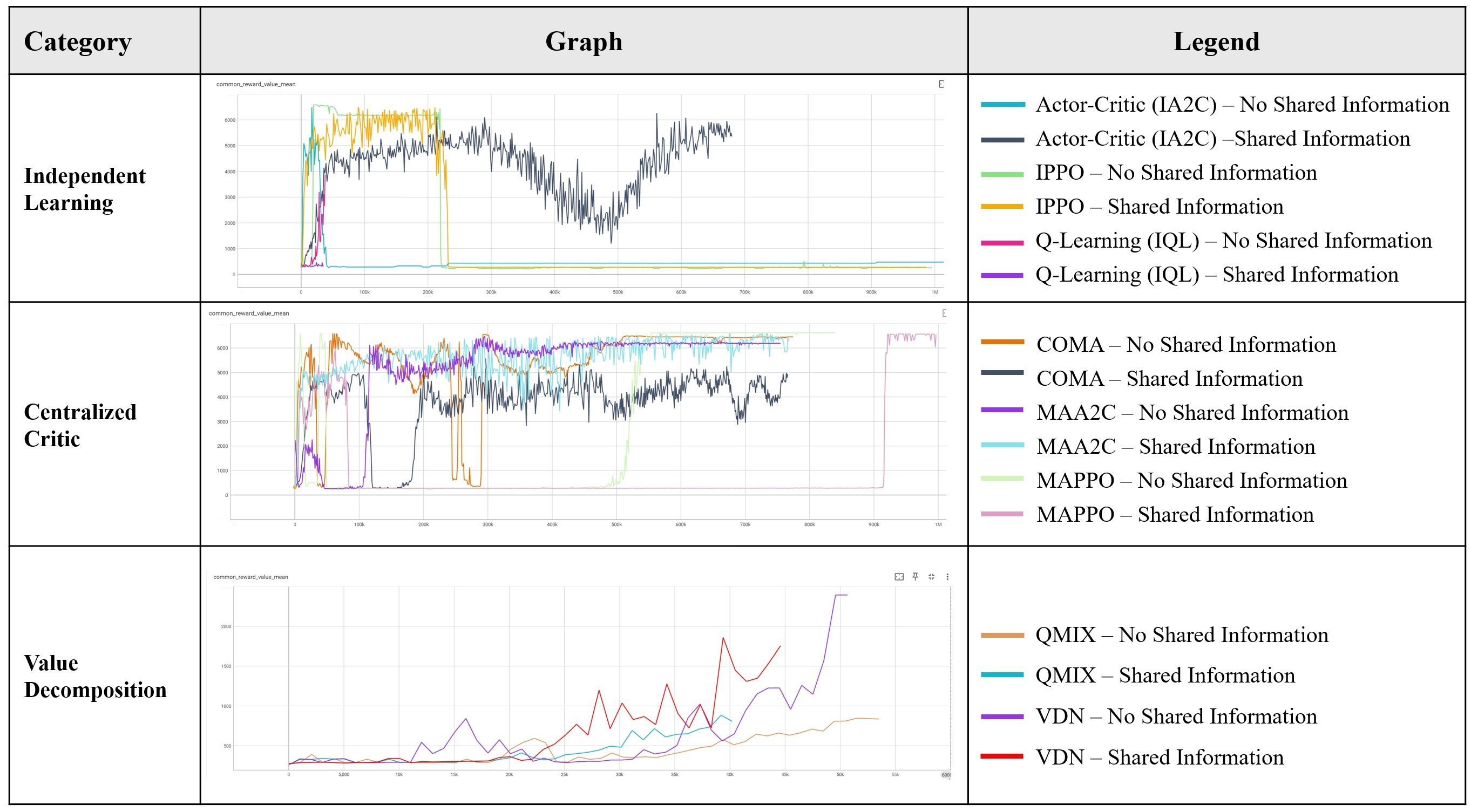}
\caption{Proof-of-concept training curves, grouped by algorithm family and information-sharing setting.}
\label{fig:6}
\end{figure*}

\textbf{Learning Algorithm:} A central challenge in multi-agent systems is that from each agent's perspective, the environment is nonstationary, and the joint state--action space grows combinatorially with the number of agents \citep{lowe2017}. Traditional optimization techniques, such as mixed-integer programming (MIP), and heuristics, including genetic algorithms and simulated annealing, can provide theoretical benchmarks under fixed assumptions but may become computationally prohibitive in large multi-agent settings. In comparison, learning-based approaches offer two key advantages: efficiency, as trained policies enable real-time decision-making without repeatedly solving costly optimization problems, making them suitable for human-in-the-loop applications; and scalability, as GPU-accelerated computation allows these methods to handle larger and more complex environments. We discuss independent learners and centralized-critic methods below; the proof-of-concept comparison also includes value decomposition (\figref{6}).

Independent learning is a family of algorithms where each agent treats other agents as part of the environment and optimizes its own return using standard RL methods such as asynchronous advantage actor--critic (A3C; \citealp{mnih2016}) and proximal policy optimization (PPO; \citealp{schulman2017}). This approach is scalable but often exhibits instability due to the nonstationary policies of other agents \citep{foerster2018,yu2021}. In our environments, independent learners tend to converge prematurely to local optima and later exhibit performance degradation; the returns do not stabilize in these runs.

With CTDE (centralized training and decentralized execution), a critic uses global state information during training, and agents act independently at execution time \citep{lowe2017}. This stabilizes value estimation and enables coordination signals to be embedded in training. In our tests, CTDE methods such as multi-agent PPO (MAPPO; \citealp{yu2021}) and multi-agent advantage actor--critic (MAA2C) generally achieved more stable convergence. Particularly, MAPPO extends the standard PPO algorithm into the multi-agent setting by combining decentralized policy updates with a centralized value function. By leveraging PPO's stability through its clipped surrogate objective, MAPPO enables multiple agents to train with a shared critic that incorporates global state information, making it well-suited for cooperative--competitive environments such as the proposed board game framework. In our experiments, MAPPO not only converged more rapidly but also sustained higher long-term returns compared to independent learners. Figure~\ref{fig:4} shows the training architecture.

\subsection{Human--Agent Interface}

We developed an interface that allows people to observe or play with trained agents (\figref{5}). The interface has two modes: simulation and play. In simulation mode, all agents are autonomously controlled by the trained MARL policies. The environment runs until a terminal state is reached, allowing researchers to evaluate emergent behaviors and collective dynamics without human intervention. We observed that increasing the emphasis on self-interest tends to correlate with higher cumulative returns for agents, suggesting that personal and collaborative reward can be aligned under certain conditions. It may also reflect insufficient competitive tension that requires further fine-tuning of the game mechanism. In Play mode, the game can be played sequentially, where humans can act as one of the agents by overwriting their actions at each step. This enables human-in-the-loop interactions, which is useful in studying hybrid decision-making, human and agent behaviors, and testing new policies or strategies.

\begin{table*}[!t]
\centering
\caption{Experiment setup. The decision-space row records the configurations as listed in the ACADIA layout; the agent, cell, and project counts are reported separately.}
\label{tab:setup}
\small
\setlength{\tabcolsep}{5pt}
\renewcommand{\arraystretch}{1.15}
\begin{tabularx}{\linewidth}{@{}>{\raggedright\arraybackslash}p{0.20\linewidth}*{3}{>{\raggedright\arraybackslash}X}@{}}
\toprule
\textbf{Parameter} & \textbf{Proof of concept} & \textbf{Scaled-up planning} & \textbf{Tatsumi, Tokyo} \\
\midrule
Decision space as reported & $3 \times 16 \times 3$ & $4 \times 44 \times 6$ & $4 \times 44 \times 6$ \\
Grid size & $4 \times 4$ & $8 \times 8$ & $12 \times 12$ \\
Duration (timesteps) & 100 & 100 & 200 \\
Agent count & 3 & 3 & 4 \\
\addlinespace
\multicolumn{4}{@{}l}{\textbf{Stakeholder weights and initial resources}} \\
Altruistic & $(0.2,0.8)$; 50/50 & $(0.2,0.8)$; 50/50 & $(0.2,0.6,0.2)$; 100/80 \\
Balanced & $(0.5,0.5)$; 50/50 & $(0.5,0.5)$; 50/50 & $(0.34,0.33,0.33)$; 90/90 \\
Interest-driven & $(0.8,0.2)$; 50/50 & $(0.8,0.2)$; 50/50 & $(0.6,0.2,0.2)$; 80/100 \\
Environment-focused & --- & --- & $(0.2,0.2,0.6)$; 80/100 \\
\addlinespace
\multicolumn{4}{@{}l}{\textbf{Map parameters}} \\
Indices (initial values, where listed)
  & Greenery 15; vitality 20; density 30
  & Sustainability 20; well-being 25; resilience 15; climate 10
  & Greenery; vitality; density; adaptability; sustainability; flood resistance \\
\addlinespace
\multicolumn{4}{@{}l}{\textbf{Projects and construction costs}} \\
Project types & 3 & 6 & 8 \\
Basic projects & Park (5/15) & House (10/5) & House (5/2) \\
 & House (10/10) & Shop (12/5) & Shop (6/3) \\
 & Shop (15/5) & & Office (7/2) \\
 & & & Factory (14/4) \\
Resilience projects & --- & Park (15/12) & Park (4/8) \\
 & & Community hub (20/15) & Shelter (5/10) \\
 & & Solar grid (25/8) & Watergate (6/8) \\
 & & Flood barrier (30/10) & Flood barrier (3/12) \\
\addlinespace
\multicolumn{4}{@{}l}{\textbf{Spatial constraints}} \\
Buildable cells & 16 & 41 & 60 \\
Non-buildable cells & None & 19 & 31 \\
Pre-built cells & None & 4 & 29 \\
\bottomrule
\end{tabularx}
\par\smallskip
\begin{minipage}{\linewidth}
\footnotesize
Decision space is expressed as players $\times$ cells $\times$ building types.
Weights are ordered as $(\alpha,\beta)$ or $(\alpha,\beta,\gamma)$.
Initial resources and construction costs are written as money/reputation.
Pre-built counts come from the source table artwork.
\end{minipage}
\end{table*}

\begin{samepage}
\section{Experiments}

We evaluated the framework in three settings: a minimal board, a larger board with resilience objectives, and a simulation initialized from a real site. Table~\ref{tab:setup} summarizes the settings.

\end{samepage}

\subsection{Proof-of-Concept Experiment}

As an initial validation, we designed a minimal-scale experiment to test whether the system can learn high-return strategies in a small multi-agent setting. The environment was configured as a $4\times4$ grid (16 cells) with a limit of 100 timesteps, involving three agents representing distinct stakeholder types: an altruistic player, a balanced player, and an interest-driven player. Each agent started with equal initial resources and pursued their objectives by balancing personal gains with collective outcomes. Players could choose among three building types---Park, House, and Shop. These actions directly affected both the environment indices---Greenery (G), Vitality (V), and Density (D)---and the agents' long-term payoffs, introducing trade-offs between short-term costs and long-term benefits. This configuration represents the minimal setup of our game: it captures the essential characteristics of multi-agent interaction and conflicting objectives across three dimensions (environmental, economic, and social), while keeping the state and action space small enough to allow clear comparison between the return from the optimization benchmark and the performance of trained models. To obtain an optimization-based benchmark, we used simulated annealing to search for high-return building placements. The search found a total return of 8,336 over the 100-timestep horizon. The reinforcement learning agents achieved total episode returns approaching 7,000, demonstrating they successfully learned strategies that capture approximately 84\% of the simulated annealing benchmark. This comparison measures performance against the best solution found by the search; it does not establish a bound on the optimal return. Among tests, MAPPO and MAA2C with shared information converged more quickly and demonstrated more stable performance than the other algorithms. Therefore, we adopt MAPPO with shared information as the training method for subsequent experiments. The training curves are shown in \figref{6}.

\begin{figure}[!htbp]
\centering
\includegraphics[width=\linewidth]{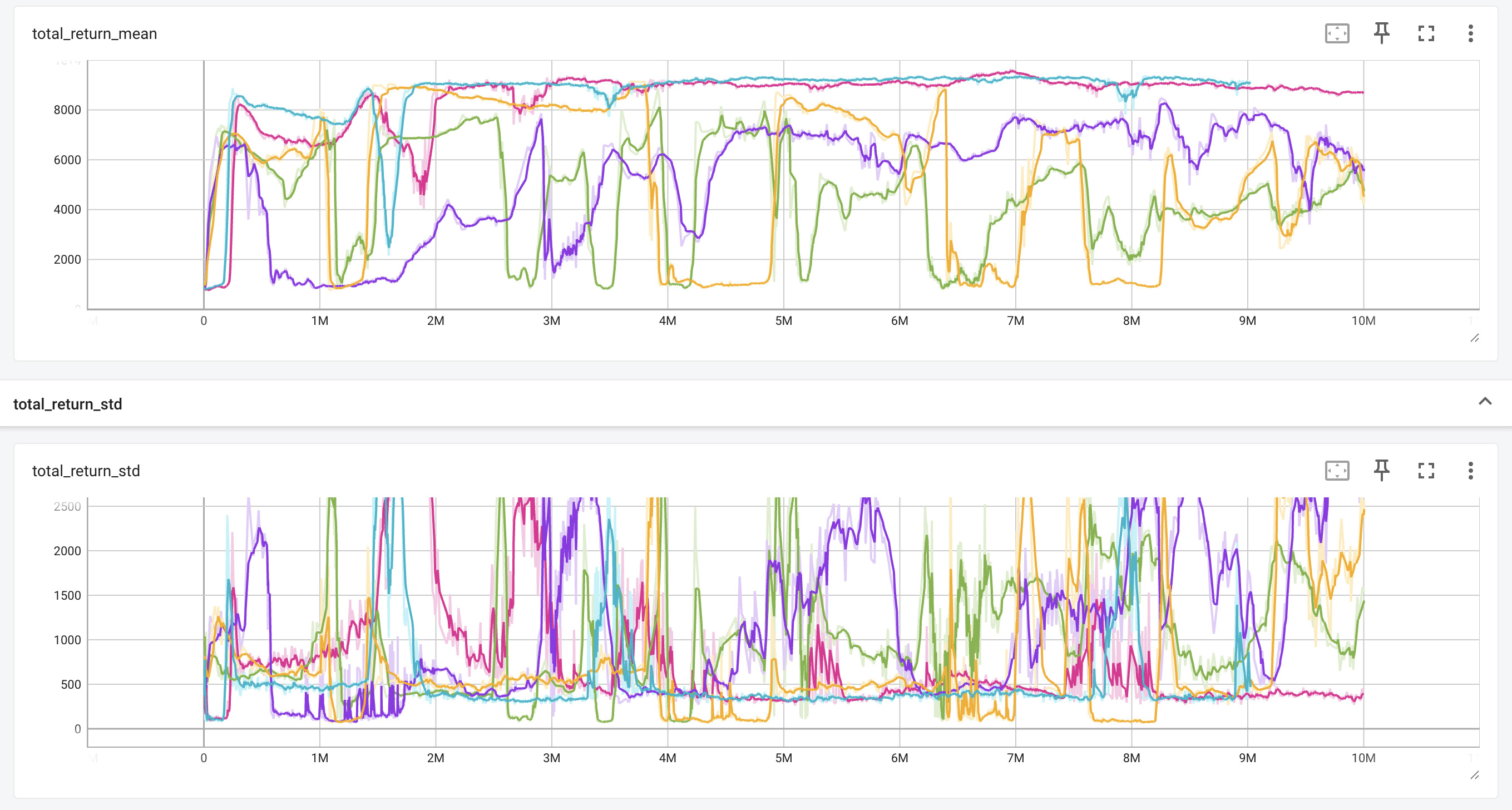}
\caption{Scaled-up MAPPO training: mean and standard deviation of total return across episodes.}
\label{fig:7}
\end{figure}

\begin{figure*}[!t]
\centering
\begin{minipage}{0.96\linewidth}
\includegraphics[width=\linewidth,trim={0bp 211bp 0bp 0bp},clip]{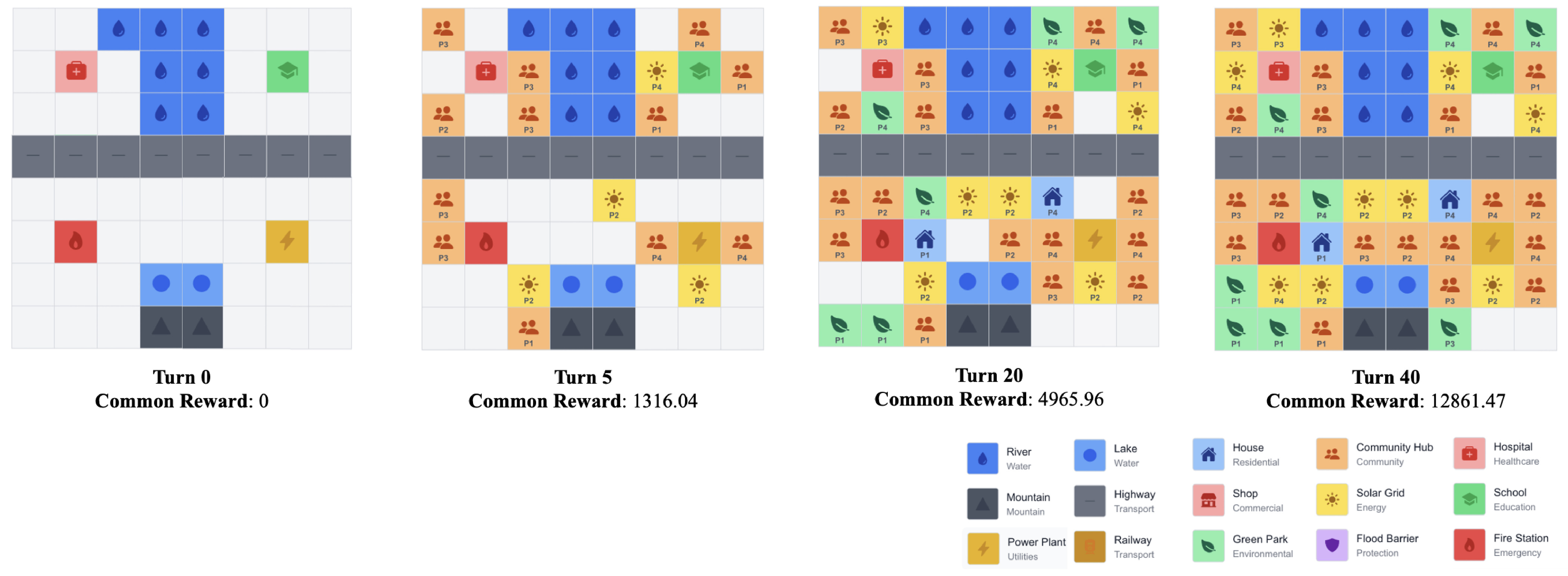}
\par\vspace{5pt}
\begingroup
\setlength{\tabcolsep}{0pt}
\renewcommand{\arraystretch}{1}
\begin{tabular*}{\linewidth}{@{\extracolsep{\fill}}*{15}{c}@{}}
\includegraphics[width=11pt,trim={1295bp 136bp 762bp 592bp},clip]{figures/fig-8.jpg} &
\includegraphics[width=11pt,trim={1438.5bp 141bp 618.5bp 587bp},clip]{figures/fig-8.jpg} &
\includegraphics[width=11pt,trim={1597.5bp 141.5bp 459.5bp 586.5bp},clip]{figures/fig-8.jpg} &
\includegraphics[width=11pt,trim={1763.5bp 142bp 293.5bp 586bp},clip]{figures/fig-8.jpg} &
\includegraphics[width=11pt,trim={1947bp 142.5bp 110bp 585.5bp},clip]{figures/fig-8.jpg} &
\includegraphics[width=11pt,trim={1295bp 75bp 762bp 653bp},clip]{figures/fig-8.jpg} &
\includegraphics[width=11pt,trim={1438.5bp 78bp 618.5bp 650bp},clip]{figures/fig-8.jpg} &
\includegraphics[width=11pt,trim={1597.5bp 80.5bp 459.5bp 647.5bp},clip]{figures/fig-8.jpg} &
\includegraphics[width=11pt,trim={1763.5bp 81bp 293.5bp 647bp},clip]{figures/fig-8.jpg} &
\includegraphics[width=11pt,trim={1947bp 81bp 110bp 647bp},clip]{figures/fig-8.jpg} &
\includegraphics[width=11pt,trim={1295bp 14.5bp 762bp 713.5bp},clip]{figures/fig-8.jpg} &
\includegraphics[width=11pt,trim={1438.5bp 17bp 618.5bp 711bp},clip]{figures/fig-8.jpg} &
\includegraphics[width=11pt,trim={1597.5bp 18bp 459.5bp 710bp},clip]{figures/fig-8.jpg} &
\includegraphics[width=11pt,trim={1763.5bp 19.5bp 293.5bp 708.5bp},clip]{figures/fig-8.jpg} &
\includegraphics[width=11pt,trim={1947bp 19.5bp 110bp 708.5bp},clip]{figures/fig-8.jpg}\\[2pt]
\shortstack{{\fontsize{5.5}{6.2}\selectfont\bfseries River}} &
\shortstack{{\fontsize{5.5}{6.2}\selectfont\bfseries Lake}} &
\shortstack{{\fontsize{5.5}{6.2}\selectfont\bfseries House}} &
\shortstack{{\fontsize{5.5}{6.2}\selectfont\bfseries Community}\\{\fontsize{5.5}{6.2}\selectfont\bfseries Hub}} &
\shortstack{{\fontsize{5.5}{6.2}\selectfont\bfseries Hospital}} &
\shortstack{{\fontsize{5.5}{6.2}\selectfont\bfseries Mountain}} &
\shortstack{{\fontsize{5.5}{6.2}\selectfont\bfseries Highway}} &
\shortstack{{\fontsize{5.5}{6.2}\selectfont\bfseries Shop}} &
\shortstack{{\fontsize{5.5}{6.2}\selectfont\bfseries Solar}\\{\fontsize{5.5}{6.2}\selectfont\bfseries Grid}} &
\shortstack{{\fontsize{5.5}{6.2}\selectfont\bfseries School}} &
\shortstack{{\fontsize{5.5}{6.2}\selectfont\bfseries Power}\\{\fontsize{5.5}{6.2}\selectfont\bfseries Plant}} &
\shortstack{{\fontsize{5.5}{6.2}\selectfont\bfseries Railway}} &
\shortstack{{\fontsize{5.5}{6.2}\selectfont\bfseries Green}\\{\fontsize{5.5}{6.2}\selectfont\bfseries Park}} &
\shortstack{{\fontsize{5.5}{6.2}\selectfont\bfseries Flood}\\{\fontsize{5.5}{6.2}\selectfont\bfseries Barrier}} &
\shortstack{{\fontsize{5.5}{6.2}\selectfont\bfseries Fire}\\{\fontsize{5.5}{6.2}\selectfont\bfseries Station}}\\[2pt]
\end{tabular*}\par
\endgroup

\end{minipage}
\caption{Board states from a scaled-up self-play session at turns 0, 5, 20, and 40.}
\label{fig:8}
\end{figure*}

\subsection{Scaled-Up Experiment}

To evaluate collaborative strategies under additional constraints and with resilience as a goal, we doubled each grid dimension to create an $8\times8$ grid, adding terrains and pre-built urban infrastructure, simulating the spatial constraints of city planning. We set the training objective to earn maximum cumulative reward under resource constraints, balancing essential demand projects with long-term resilience investments. For this experiment, we conducted five training runs with different random seeds, each for 10 million episodes. In \figref{7}, the top plot shows the mean total return over a training session on the scaled-up environment using the MAPPO algorithm; the bottom plot tracks the standard deviation of returns across episodes. Despite the instability of built-in multi-agent competition, the overall return generally increases. We observed several inflection points, indicating policy shifts as agents discover new coordination strategies. For some training sessions, the return curve stabilizes near its peak, suggesting convergence to a high-reward collaborative policy. The declining standard deviation further reflects improved policy stability and reduced variance across episodes.

A simulated gameplay session is visualized in \figref{8}, captured from the human--agent interface. In this session, resilience-oriented projects, including community hubs, parks, and solar grids, occupied 82\% of the buildable cells. These results reveal that, under dynamic spatial and resource trade-offs, the trained strategies evolve through self-play to internalize spatial--reward relationships between project types and environmental outcomes. The agents ultimately form adaptive and context-sensitive strategies that optimize collaborative decisions in complex, uncertain urban scenarios, which can be further used as a dynamic reference in real-world zoning and resource allocation practices under the theme of resilience.

\begin{figure}[!htbp]
\centering
\includegraphics[width=\linewidth,trim={32.5bp 410bp 23.5bp 97.5bp},clip]{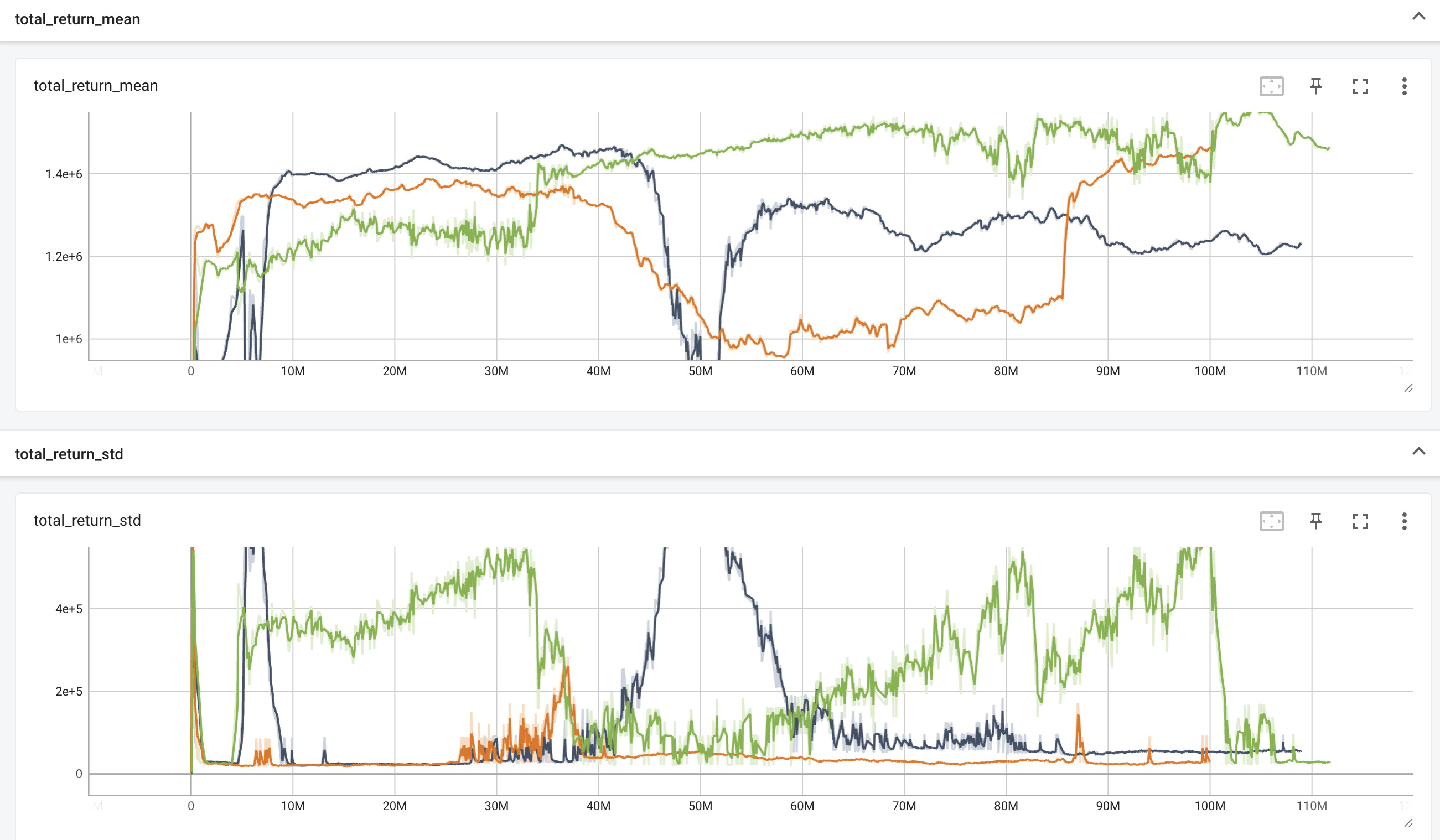}
\par\vspace{2pt}
{\raggedright\footnotesize Total Return Mean\par}
\vspace{8pt}
\includegraphics[width=\linewidth,trim={32.5bp 9bp 23.5bp 492.5bp},clip]{figures/fig-9.jpg}
\par\vspace{2pt}
{\raggedright\footnotesize Total Return Standard Deviation\par}
\caption{Tatsumi MAPPO training: mean and standard deviation of total return for three runs.}
\label{fig:9}
\end{figure}

\begin{figure*}[!t]
\centering
\includegraphics[width=\linewidth]{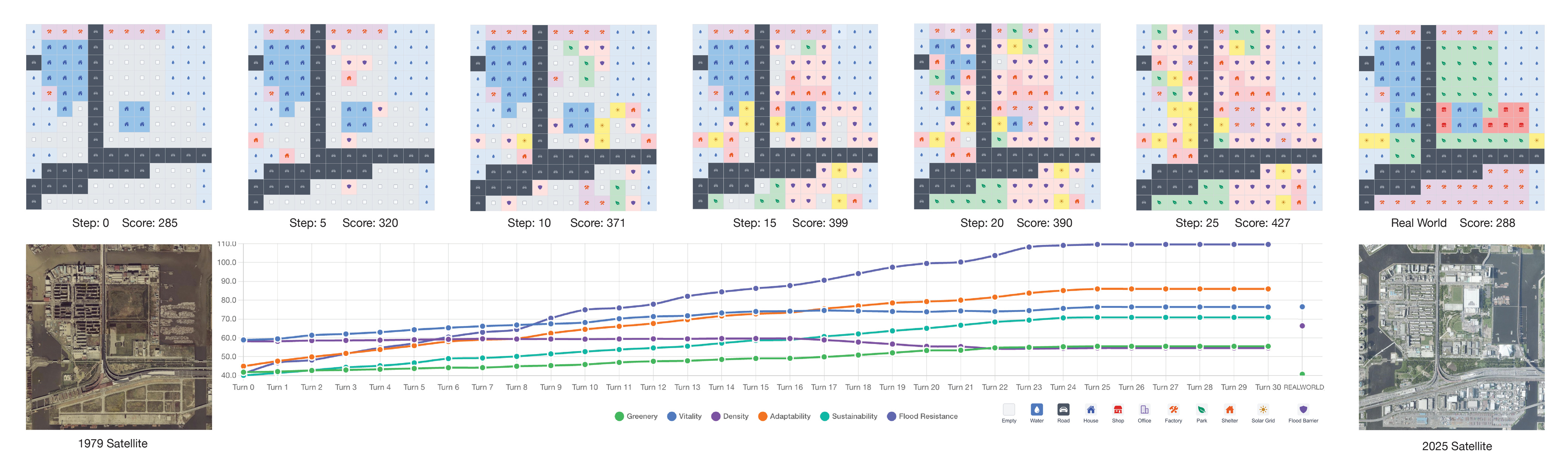}
\caption{Tatsumi self-play: board states and scores over time, with the 2025 map as a comparison. Source of aerial photographs (1979--1983 and 2025): \href{https://maps.gsi.go.jp/development/ichiran.html}{Geospatial Information Authority of Japan (GSI)}. Images cropped and annotated by the author.}
\label{fig:10}
\end{figure*}

\subsection{Tatsumi, Tokyo Case Study}

Based on the previous experiments, we extend our model to a real-world site to demonstrate its adaptability and scalability. For this purpose, we selected a district in Koto City, Tokyo. The area is built on low-lying wetlands and frequently exposed to typhoons, making it a critical site for examining historical needs for resilience improvement. However, due to resource constraints, the district requires careful management of resource allocation under stakeholders' preferences. We selected the 1980s as the initial period, which marks the beginning of the peak of urban development and can help us compare simulated results with real-world development results. We initialize the $12\times12$ grid directly from historical maps: for each cell, we extract the dominant parcel or land-cover feature and assign it to the closest in-game type---either a terrain class or a pre-built project. These assignments are treated equivalently to in-game constructions: they apply the same local and $3\times3$ neighborhood effects, incur the same per-turn costs/benefits, and thereby initialize the full set of map parameters at the start of each episode. This larger decision space makes exhaustive search less practical, so we use learned policies to explore the available strategies. Following the previous experiment, we adopt MAPPO, which was shown to be effective, as the training algorithm in this setting. We trained agents in the environment for 100 million episodes with three random seeds, as shown in \figref{9}. Overall, the training process exhibits high volatility in the early stage as agents rapidly explore the environment. After converging to several local optima, some runs (e.g., the green curve) continue to improve and eventually achieve higher returns. \figref{10} shows a typical simulated game where the total score rises overall from 285 to 427 points, about 48\% above the 2025 map baseline of 288 when evaluated with the same scoring model. We further observe that resilience-related scores and growth rates gradually surpass those of non-resilience projects, indicating that the resulting policies increasingly favor resilience-oriented investments, which, despite their higher initial costs, yield greater long-term rewards.

\begin{figure*}[!t]
\centering
\includegraphics[width=0.56\linewidth]{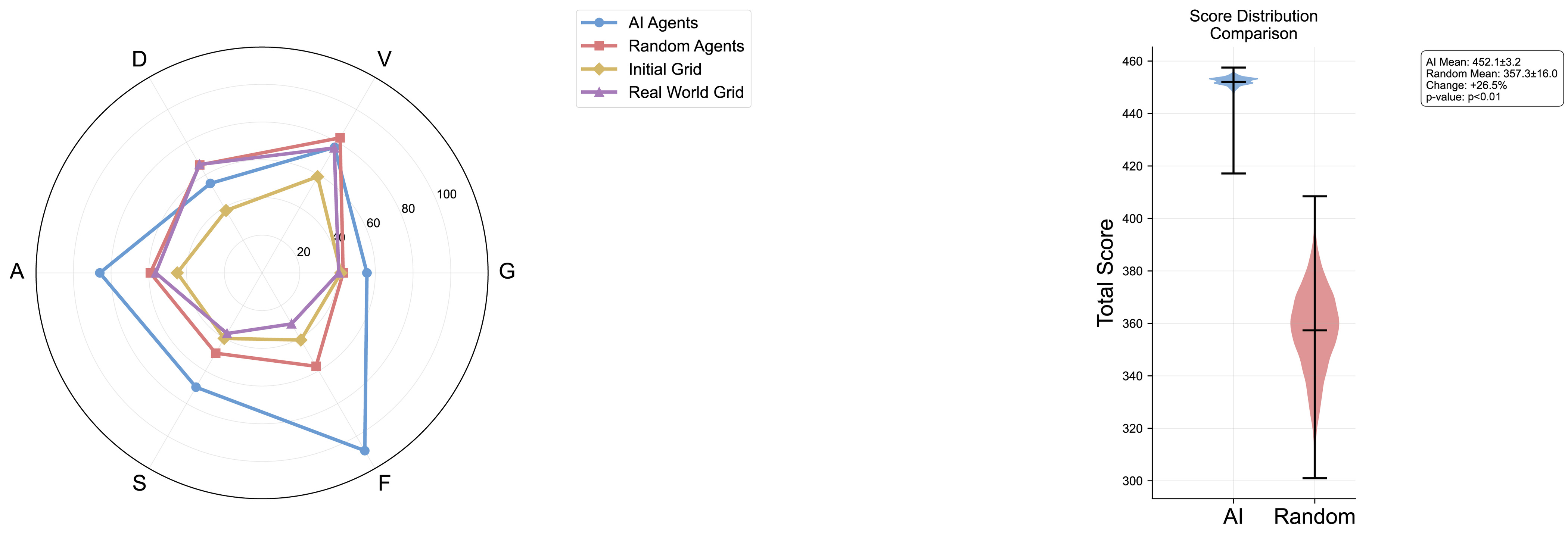}\hfill
\includegraphics[width=0.41\linewidth]{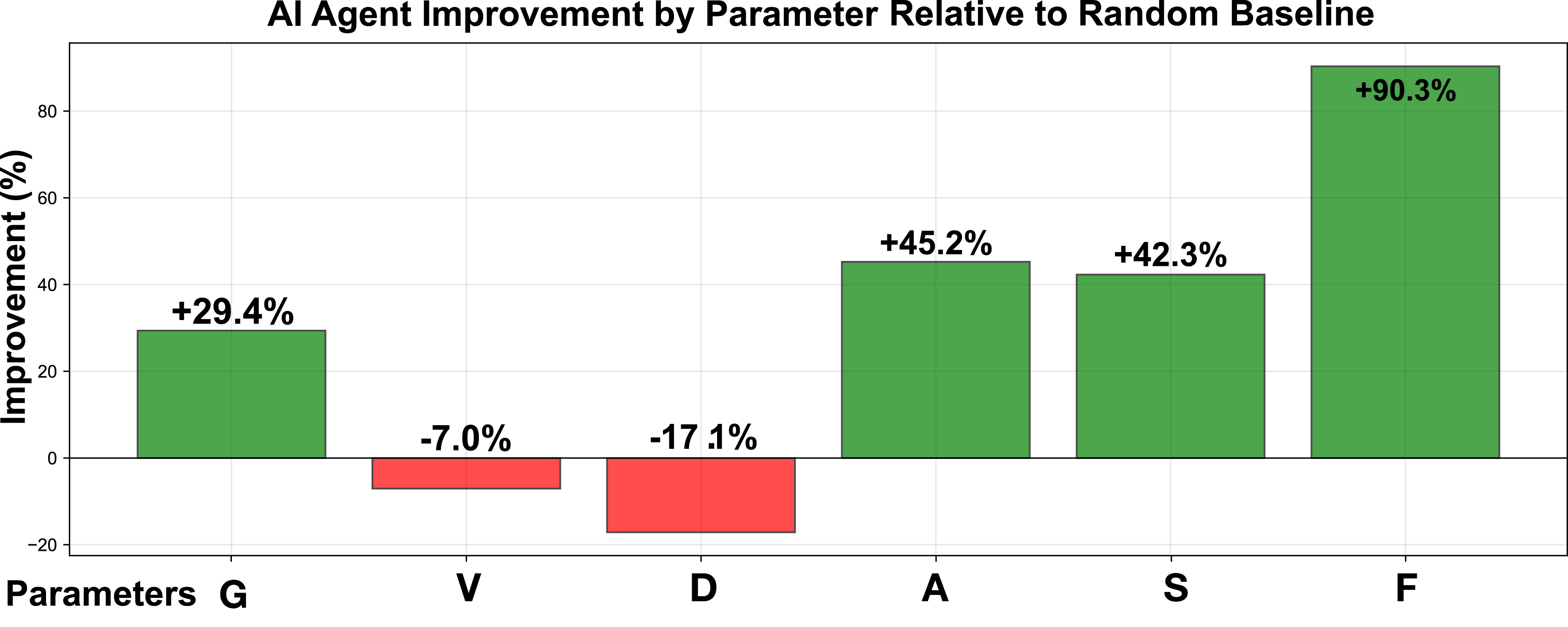}
\caption{Tatsumi comparison across 10,000 simulation runs: parameter profiles and score distributions (left), and relative changes in each parameter versus the random-policy baseline (right).}
\label{fig:11}
\end{figure*}

To compare the trained and random policies, we conducted 10,000 independent simulation runs (\figref{11}). Compared with the random baseline, the AI agents achieved a higher average total score (452.1, standard deviation (SD) = 3.2 vs.~357.3, SD = 16.0; $p<0.01$) and a more concentrated distribution (interquartile range (IQR): 1.82 vs.~21.19). The trained policies scored higher on Adaptability (A), Sustainability (S), Flood Resistance (F), and Greenery (G), and lower on Vitality (V) and Density (D). These differences show how the policies trade off resilience and development under the specified scoring model.

\section{Conclusion and Future Work}

This study presents a practical framework for modeling and evaluating resource allocation processes as a trainable digital board game, offering a novel bridge between game theory, multi-agent learning, and participatory planning. Through mechanism design, the framework captures the complex trade-offs between individual incentives and collective outcomes in support of urban resilience. Notably, our scaled-up simulation shows that the resulting policies consistently emphasize resilience-oriented projects while maintaining overall urban viability through decentralized self-play, forming context-aware, adaptive strategies. The gamified environment serves not only as a learning environment for agents but also as an interface for human engagement, supporting human-in-the-loop experimentation, which builds a foundation for explainable, interactive simulation with a focus on transforming how urban policy is prototyped and negotiated in future AI-augmented decision-making.

Several limitations remain. First, the reliance on an orthogonal grid restricts the accurate representation of irregular urban block structures. Second, the scoring system is based on predefined objective indices, which may oversimplify or misrepresent the nuanced trade-offs inherent in real-world decision-making. Third, reinforcement learning depends heavily on trial-and-error exploration, resulting in relatively low training efficiency and potential instability or convergence challenges when scaling to larger environments and action spaces. We leave formal equilibrium tests for future work. Finally, although the framework provides an interactive interface that allows humans to observe and engage with the decision-making process, it has not yet been evaluated through full-scale human participant studies.

Looking ahead, future work will focus on scaling the environment and agent population to support more complex and realistic conditions. Another direction is to integrate large language models to support human-interpretable reasoning processes and structured chains of thought, incorporating real-life common sense and its influence on complex decision-making issues. Last but not least, the framework will be applied to real-world planning scenarios with human participants to explore its applicability in an AI-driven, participatory design process.

\section*{Acknowledgments}

The author wishes to thank Professor Gabriele Farina, Zhiyuan Fan, Professor Miho Mazereeuw and the MIT Urban Risk Lab, Professor Marcelo Coelho and Professor Manish Raghavan for their generous advice, encouragement, and valuable discussions that greatly contributed to this work.

\section*{Image Credits}
Aerial photographs in \figref{10}: Geospatial Information Authority of Japan (GSI),
Ministry of Land, Infrastructure, Transport and Tourism. The author cropped and annotated the imagery.
The training diagram in \figref{4} is adapted from \citet{hu2023} (\href{https://creativecommons.org/licenses/by/4.0/}{CC BY 4.0}).
All other drawings and images are by the author.

\end{document}